\documentclass[a4paper,conference]{IEEEtran}
\IEEEoverridecommandlockouts

\usepackage{url}
\usepackage{paralist}
\usepackage{amsmath}
\usepackage{amssymb}

\usepackage{graphics} 

\title{\LARGE \bf
MTStereo 2.0: improved accuracy of stereo depth estimation with Max-trees
}

\author{Rafa\"{e}l Brandt, Nicola Strisciuglio and Nicolai Petkov
\thanks{Rafa\"{e}l Brandt and Nicolai Petkov are with the Bernoulli Institute, University of Groningen, The Netherlands. {\tt\small r.brandt@rug.nl}}%
\thanks{Nicola Strisciuglio is with the Faculty of Electrical Engineering, Mathematics and Computer Science at University of Twente, Netherlands.}%
\thanks{This work received support from the European Horizon 2020 program, under the project TrimBot2020 (grant No. 688007)}
}

\graphicspath{ {./Figures/} }
\usepackage[demo]{graphicx}
\usepackage{xcolor}
\usepackage{tabularx}
\usepackage{booktabs}

\usepackage{subfig}

\usepackage{comment}

\begin{document}

\maketitle
\thispagestyle{empty}
\pagestyle{empty}

\begin{abstract}
Efficient yet accurate extraction of depth from stereo image pairs is required by systems with low power resources, such as robotics and embedded systems. State-of-the-art stereo matching methods based on convolutional neural networks require intensive computations on GPUs and are difficult to deploy on embedded systems. In this paper, we propose a stereo matching method, called MTStereo 2.0, for limited-resource systems that require efficient and accurate depth estimation. It is based on a Max-tree hierarchical representation of image pairs, which we use to identify matching regions along image scan-lines. The method includes a cost function that considers similarity of region contextual information based on the Max-trees and a disparity border preserving cost aggregation approach. MTStereo 2.0 improves on its predecessor MTStereo 1.0 as it
\begin{inparaenum}[\itshape a\upshape)]
    \item deploys a more robust cost function,
    \item performs more thorough detection of incorrect matches,
    \item computes disparity maps with pixel-level rather than node-level precision.
\end{inparaenum}
MTStereo provides accurate sparse and semi-dense depth estimation and does not require intensive GPU computations like methods based on CNNs. Thus it can run on embedded and robotics devices with low-power requirements. 
We tested the proposed approach on several benchmark data sets, namely KITTI 2015, Driving, FlyingThings3D, Middlebury 2014, Monkaa and the TrimBot2020 garden data sets, and achieved competitive accuracy and efficiency. The code is available at \url{https://github.com/rbrandt1/MaxTreeS}.
\end{abstract}

\section{INTRODUCTION}
Estimation of scene depth from stereo image pairs is deployed as a building block in high level computer vision applications, such as autonomous car driving~\cite{geiger20133d,ros2015vision}, obstacle avoidance by robots~\cite{oleynikova2015reactive,schmid2013stereo}, simultaneous localization and mapping~\cite{engel2015large}, among others. 
In two-view stereo matching, the three-dimensional structure of a scene is recovered by finding correspondent pixels in image pairs. Two pixels in the left and right image, respectively $(x,y)$ and $(x-d,y)$, match when they capture the same scene point. The horizontal offset (i.e. disparity) $d$ of the pixels is used to compute the distance $z$ of the captured scene point as $z = Bf/d$, where $B$ is the camera baseline and $f$ the camera focal length. 

The similarity of two pixels is quantitatively computed by a matching cost function, 
e.g. absolute image gradient or gray-level difference~\cite{scharstein2002taxonomy}. Substantial matching ambiguity is however caused by repetitive patterns and uniformly colored regions. Hence,
costs are aggregated over neighbor pixels to strengthen the robustness of the matching evaluation. 
For instance, 
color similarity and proximity were used for disparity estimation in~\cite{yoon2006adaptive} and~\cite{zhang2009cross}. A scheme that takes into account the strength of image boundaries between pixels was proposed in~\cite{chen2013improved}. 
Early methods performed exhaustive matching search~\cite{arnold1983automated}, which requires many computations. Later approaches reduced the disparity search-range by computing a coarse disparity map first and then refining it in an iterative approach~\cite{geiger2010efficient}. Image pyramids were also used to reduce disparity search range in~\cite{sun1997fast,luo2015fast}. A coarse 
disparity map is estimated considering the full disparity range. Then, increasingly higher-resolution disparity maps are constructed whereby the disparity search range is dictated by the previously computed (coarser) disparity map. To increase efficiency and reduce matching ambiguity, matching (hierarchically structured) image regions instead of individual pixels was proposed~\cite{cohen1989hierarchical,medioni1985segment,todorovic2008region}. Such methods may require computationally expensive segmentation steps.

Recent approaches use convolutional neural networks (CNNs) to compute aggregated matching costs.
One of the first CNN-based stereo matching methods deployed a siamese network architecture~\cite{zbontar2016stereo}, which works with small patch inputs. 
Approaches have been suggested to increase the receptive field while maintaining details in estimated disparity maps. In~\cite{chen2015deep}, pairs of siamese networks, each receiving as input a pair of patches at different scales were used and the matching cost was computed as the inner product between the responses of the siamese networks. 
Various approaches were further developed to improve the quality and accuracy of depth maps estimated by CNNs, namely recurrent architecture blocks~\cite{cheng2018learning}, stacked modules with short-cut connections and separable convolutions~\cite{1904.09099}, group-wise convolutions~\cite{guo2019group}, 3D convolutional layers as in GC-Net~\cite{kendall2017end}, and layers focused on local- and whole-image features to compute cost dependencies~\cite{zhang2019GANet}.

Although CNN-based methods compute highly accurate disparity maps, 
they need power-consuming dedicated hardware or GPUs to compute the many convolutions. This limits their usability on embedded or power-constrained systems.  This applies, for instance, to battery-powered robots or drones, for which depth estimation has to trade-off between accuracy and computational efficiency. Furthermore, for robot navigation and obstacle avoidance, the very high accuracy and density of estimation achieved by CNNs are not strictly necessary. 

In this paper, we present a stereo matching method, named MTStereo 2.0, that balances efficiency with effectiveness, making it appropriate for devices where limited computational and energy resources are available. The MTStereo 1.0 algorithm, that we proposed in~\cite{BRANDT2020}, exploits contrast information of objects in a hierarchical fashion to efficiently and effectively perform stereo matching. It constructs a hierarchical representation of image scan-lines using  Max-Trees~\cite{salembier1998antiextensive}, and performs disparity estimation via tree matching cost computation that takes into account contextual image structural information. The MTStereo 2.0 algorithm that we propose improves on the 1.0 version as it 
\begin{inparaenum}[\itshape a\upshape)]
    \item deploys a more robust cost function, 
    \item performs more thorough incorrect match detection, 
    \item computes disparity maps with pixel-level rather than node-level precision.
\end{inparaenum}

We carried out an extensive experimental benchmark on several data sets, namely KITTI 2015, Driving, FlyingThings3D, Middlebury 2014, Monkaa and the TrimBot2020 garden data sets.   

\section{MTStereo 2.0}\label{sec:descriptionMethod}

We present MTStereo 2.0, which exploits contrast information of objects in a hierarchical fashion to efficiently and effectively perform stereo matching. 
In the following, we outline the elements of MTStereo 1.0 that are also present in MTStereo 2.0, and detail the novel elements that we introduced in the new version of the algorithm.
We provide an implementation of MTStereo 2.0 available at \url{https://github.com/rbrandt1/MaxTreeS}. 

\subsection{The method}
\begin{figure}[!t]
    \centering
    \includegraphics[width=.46\textwidth]{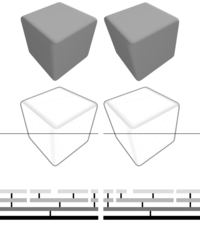}
    \caption{A (top row) pair of stereo images and their (middle row) version resulted by edge-detection (darker regions contain more contrast). The Max-Tree representations of the horizontal scan-lines highlighted in the middle row are illustrated in the below row: MTstereo 2.0 computes the disparity for coarse structure first (nodes represented by darker bars) and increasingly matches only finer structures (with lighter color) contained in earlier matched coarser ones. 
    }
    \label{fig:overview}
\end{figure}

A pair of stereo images depicting a cube is shown in Fig.~\ref{fig:overview} (top row) together with their processed version by using an edge-detector filter (middle row). Darker regions in the processed images (middle row) correspond to regions with higher contrast, while lighter regions correspond to areas with less contrast. We use the contrast information to construct Max-Trees of the image scan-lines.  In Fig.~\ref{fig:overview}, examples of Max-Trees constructed for the highlighted scan-lines in the middle row are depicted in the third row.
Matching the finest structures directly results in non-efficient yet precise stereo matching, while matching coarse structures results in disparity maps that lack precision but can be efficiently obtained. To tackle this trade-off and efficiently obtain precise disparity maps, our method matches increasingly finer regions in an iterative manner, and only compares regions contained in earlier matched coarser ones.
MTStereo 2.0 is composed of the following steps: pre-processing, Max-Tree construction, cost volume computation, cost aggregation, consistency check, disparity map computation, confidence check and map refinement.

\begin{figure*}[!ht]
    \centering
    \setlength{\unitlength}{22mm}
    \subfloat[]{\label{fig:I1}
        \includegraphics[clip,height=\unitlength]{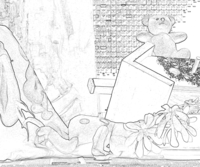}}~
    \subfloat[]{\label{fig:I2}
        \includegraphics[clip,height=\unitlength]{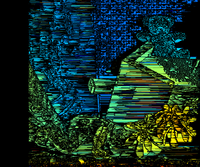}}~
    \subfloat[]{\label{fig:I3}
        \includegraphics[clip,height=\unitlength]{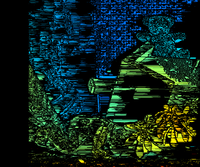}}~
    \subfloat[]{\label{fig:I4}
        \includegraphics[clip,height=\unitlength]{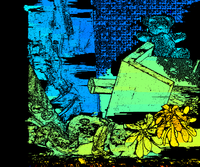}}~ 
    \subfloat[]{\label{fig:I5}
        \includegraphics[clip,height=\unitlength]{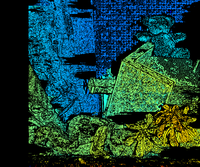}}~
    \subfloat[]{\label{fig:I6}
        \includegraphics[clip,height=\unitlength]{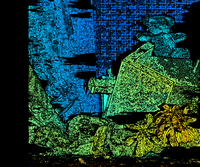}}
    \caption{Outputs at intermedate algorithm stages: (a) pre-processed image of Middlebury training data set, (b) output of coarse-to-fine matching (c) maps after outlier removal, (d) reliable node extrapolation, (e) output of guided pixel matching, (f) output map after outlier removal.}
    \label{fig:methodOverview}
\end{figure*}

\paragraph{Pre-processing}
We process the input rectified image pair images with a median filter for noise removal.  
%
Subsequently, we detect edges in the images by a horizontal and a vertical Sobel operator, and average their absolute response images pixel-wise. We invert the result image, perform contrast-stretch and color quantization. We show an example of processed image in Fig.~\ref{fig:methodOverview}a. We compute a 1D Max-Tree for each scan-line of the pre-processed images. A parameter $q$ controls the number of colors for color quantization and influences the size of the constructed trees. Shallower trees are less expensive to match, but represent less precisely the image structures. 

\paragraph{Max-Tree construction}
The coarse-to-fine matching is facilitated by a hierarchical representation of both images in a rectified stereo pair: we compute 1D Max-Trees on the scan-lines the images. We store regions with less contrast being contained in regions with more contrast~\cite{todorovic2008region}. The Max-Tree, proposed by~\cite{salembier1998antiextensive}, allows storing the hierarchy of connected components resulting from different thresholds and is efficiently constructed.  A 1D connected component set contains the 1-valued pixels for which no 0-valued pixel exists in between any of the pixels in the binary image resulting from applying a threshold $t$ to a 1D gray-scale image (i.e. a scan-line). We construct 1D Max-Trees using the algorithm in~\cite{wilkinson2011fast}. 

\paragraph{Cost volume}
We construct a cost volume by computing the matching cost of each pixel in the left image with those in the right image at all possible disparity levels. The matching cost of two pixels is the weighted-average of the absolute difference between their gray-level, horizontal Sobel, and vertical Sobel values. We process each slide of the cost volume with a Gaussian blur operator, which smooths the estimation in textured areas and facilitates the subsequent steps of the algorithm. The parameter $\omega_{cv}$ controls the size of the Gaussian blur kernel ($\omega_{cv} \times \omega_{cv}$). MTStereo 1.0 does not make use of a smoothed cost volume and does not consider gray-level difference in the cost computation.

\paragraph{Matching cost aggregation}
The constructed cost volume and tree structures are then used for matching of regions in image pairs. 
The nodes corresponding to coarse image structures are initially matched. Thereafter, increasingly finer structures are matched. Only nodes with a coarseness in set $S$ are matched (see \cite{BRANDT2020} for details). Only the finest nodes and their descendants are matched, of which the width is greater than $\omega_\alpha$ and less than $\omega_\beta$. 

The matching cost of node pairs consists of a context cost and an intensity cost. The context cost is the average relative difference between the area of corresponding ancestors of a node pair. Given a node pair ($n_1$,$n_2$), we define that its ancestors ($n_3$,$n_4$), where $n_3$ is an ancestor of $n_1$ and $n_4$ is an ancestor of $n_2$, can be matched if in the Max-Trees the number of nodes between $n_3$ and $n_1$, and $n_4$ and $n_2$ is equal. We define the intensity cost as the average of the cost volume matching costs at aligned pixels part of the two nodes. Given a node pair, we compute the matching cost at the location of the left node's left (right) endpoint and disparity between the left (right) endpoints of both nodes, as well as at linearly interpolated disparity and pixel values in between the two endpoints. This definition of intensity cost is more robust than that of MTStereo 1.0. Parameter $\alpha$ controls the relative weight of the intensity cost ($\alpha$) and context cost ($1-\alpha$). 

Let $P_y = (n_{L,y}, n_{R,y})$ denote a pair of nodes, respectively in the left and right image, both at row $y$. Matching cost is aggregated over their node neighborhood. We define the neighborhood of $P_y$ as the nodes with the same coarseness level as $P_y$ that have similar x-coordinates and are in scan-lines next to each other in the original image. Two nodes have the same coarseness level when the distance (i.e. the difference in tree level) between the leaf nodes with the greatest level out of the leaf nodes which are descendants of the nodes, and the nodes themselves are equal. Recursively, node pair $P_{y+1}$ is part of the neighborhood of $P_{y}$, if both $n_{L,y+1}$ crosses the x-coordinate of the center of node $n_{L,y}$, $n_{R,y+1}$ crosses the x-coordinate of the center of node $n_{R,y}$, and both $n_{L,y+1},$ and $n_{R,y+1}$ have a y-coordinate which is one lower or higher than that of $(n_{L,y}$ and $n_{R,y})$. At most, the $\omega_\gamma$ nodes above and below a node pair are included in the neighborhood of $P_y$. In the coarse-to-fine matching procedure, we compute a disparity search range for each node in each iteration. Given a node pair that has likely been correctly matched in a previous iteration, only descendants of this node pair are matched in subsequent iterations. Nodes are considered likely correctly matched when they pass a left-right consistency check~\cite{weng1988two} and a confidence check (peak-ratio used in~\cite{yang2014fast}). The confidence check, not used in MTStereo 1.0, provides more accurate disparity maps as it filters out ambiguous (and thus likely incorrect) matches.

\begin{table*}[!t]
  \centering
  \renewcommand{\arraystretch}{1.3}
\caption{Details of the data sets used for the experiments.}\label{tab:datasets}
  \begin{tabular}{p{3cm}p{4cm}cp{2cm}p{3.5cm}c}
  \toprule
\textbf{Dataset} & \textbf{Description} & \textbf{Synthetic} & \textbf{Size in pixels} & \textbf{Subset} & \textbf{Dense GT} \\

 \cmidrule(lr){1-1}  \cmidrule(lr){2-2}     \cmidrule(lr){3-3}     \cmidrule(lr){4-4}     \cmidrule(lr){5-5}     \cmidrule(lr){6-6}%

Middlebury 2014~\cite{hirschmuller2007evaluation}& Indoor scenes. While average accuracy results were weighted by the official weights, average density results were not weighted. & $\times$ &  $\leq 3000~\times~2000$ & The Full-size version was used in our experiments. Some of the results which were taken from the benchmark website were generated using a down-scaled version of the data set. & $\checkmark$  \\ 

Kitti 2015~\cite{Menze2015ISA}& Street scenes captured from the viewpoint of a driving car. & $\times$ & $\leq 1242~\times~376$ &  & $\times$\\

Trimbot2020 Synthetic~Garden~\cite{Tylecek2018rms}& Outdoor garden scenes rendered from 3D synthetic models of gardens in the context of the TrimBot2020 project~\cite{StrisciuglioTrimbot}. & $\checkmark$ & $640 \times 480$  & Test set used by \cite{pu2019sdf}. & $\checkmark$ \\

Trimbot2020 Real~Garden~\cite{sattler20173d}& Outdoor garden scenes recorded in the test garden of the said TrimBot2020 project. & $\times$ & $752 \times 480$  & Test set used by \cite{pu2019sdf}. & $\checkmark$  \\

Driving~\cite{MIFDB16} & Realistic street scene images captured from the viewpoint of a driving car. & $\checkmark$ & $960 \times 540$  & Cleanpass, fast, 35mm\_focallength & $\checkmark$\\

Monkaa~\cite{MIFDB16} & Un-naturalistic images of furry objects in outdoor scenes. & $\checkmark$ & $960 \times 540$  & Cleanpass & $\checkmark$  \\

Flying Things 3D~\cite{MIFDB16}& Textured objects moving in random 3D paths. & $\checkmark$ & $960 \times 540$  & Cleanpass, test set. Stereo pairs excluded by \cite{mayer2016large} were excluded in our experiments as well. & $\checkmark$ \\
 \bottomrule
\end{tabular}
\end{table*}

\paragraph{Disparity map}
The coarse to fine matching procedure results in a list of matched node pairs. For each of the finest nodes, the disparity at the left and right endpoints of matched nodes is computed and linearly interpolated. A resulting disparity map is illustrated in Fig.~\ref{fig:methodOverview}b.
We process the disparity map to remove outliers: a disparity value is removed when in the $(2 \cdot 21) \times (2 \cdot 21)$ local neighborhood the number of pixels with a disparity difference that exceeds their $x$-offset surpasses the number of pixels with a disparity difference less than or equal to their $x$-offset. A disparity map after noise removal is illustrated in Fig.~\ref{fig:methodOverview}c.

We improve the density and accuracy of a disparity map by computing median disparity values of nodes which are neighbors across scan-lines. Hence, nodes previously without disparity assignment obtain a disparity value and nodes with an outlier disparity assignment are corrected. The disparity values at the endpoints of the finest nodes are stored in the Max-Trees. Thereafter, for each of the finest nodes, the median of the left and right side disparity among the $\omega_\gamma$ neighbor nodes above and $\omega_\gamma$ below the node which have a disparity assigned to their endpoints are stored in the node. 
When a semi-dense disparity map is generated, the disparity at the left and right endpoints of matched nodes is linearly interpolated, while the disparity at the left and right endpoints of matched nodes is only assigned to the endpoints of the node pair if a sparse disparity map is being generated.
A disparity map after reliable node extrapolation is illustrated in Fig.~\ref{fig:methodOverview}d.
Differently from MTStereo 1.0, we perform pixel matching to recover surface shape. Disparity search range for pixels is set such that only disparities which are $\omega_\omega$ percent more or less than the previously computed disparity value are considered. The matching cost of pixel pairs is derived from the constructed cost volume. 

\paragraph{Confidence check and map refinement}
We introduce a step of confidence check on the estimated pixel disparities. If a match does not pass the confidence check, it is not taken into account for the disparity map.
A node pair passes the confidence check when the relative difference between the matching cost of the best match and that of the second-best match is more than a percentage controlled by parameter $\omega_\Pi$. Areas which contain more texture are more likely to pass the confidence check.
A disparity map after guided pixel matching is illustrated in Fig.~\ref{fig:methodOverview}e.
We obtain the final disparity map by removing the outliers. A disparity map after noise removal is illustrated in Fig.~\ref{fig:methodOverview}f.

\section{EXPERIMENTS}
We carried out an extensive evaluation of the MTStereo 2.0 performance on several benchmark data sets, of which we report details in Table~\ref{tab:datasets}, and compared its results with those of other methods. We ran our algorithm on an Intel{\small\textregistered~} Core{\small\texttrademark~} i7-2600K CPU @3.40GHz. 

\begin{table*}[!t]
  \centering
\renewcommand{\arraystretch}{1.3}  
 \setlength{\tabcolsep}{3pt}
  \caption{Results expressed in \textit{avgerr}, except Kitty2015 Test which are expressed in \textit{D-all-est}. Average prediction density is reported in brackets. Entries without density specification are 100\% dense. Entries with ?\% density specification have unknown density. We mark with $^*$ the results that were (possibly) computed on a different set of images than we used for evaluation. We mark with $^+$ that result was taken from respective benchmark website. We mark with $^T$ results that were (possibly) computed using a different metric computation approach.}
  \label{tab:accuracyResults}
\setlength{\tabcolsep}{2pt}

  \begin{tabular}{rccccccccc}
 
    \toprule
     &\multicolumn{2}{c}{\textbf{Middlebury}} &\multicolumn{2}{c}{\textbf{Kitti2015}} 
      &\multicolumn{1}{c}{\textbf{Real Garden}} &\multicolumn{1}{c}{\textbf{Synth Garden}} 
       &\multicolumn{1}{c}{\textbf{Driving}} &\multicolumn{1}{c}{\textbf{Monkaa}}
       &\multicolumn{1}{c}{\textbf{Flying Things}} \\
       
     &\multicolumn{1}{c}{Train} &\multicolumn{1}{c}{Test}  &\multicolumn{1}{c}{Train} &\multicolumn{1}{c}{Test} &\multicolumn{1}{c}{}  
     &\multicolumn{1}{c}{}  
     &\multicolumn{1}{c}{} 
     &\multicolumn{1}{c}{} 
     &\multicolumn{1}{c}{} \\
     
    \cmidrule(lr){2-3}     \cmidrule(lr){4-5}           \cmidrule(lr){6-6}     \cmidrule(lr){7-7}     \cmidrule(lr){8-8} 		
    \cmidrule(lr){9-9}     \cmidrule(lr){10-10} 

\textbf{MTS2.0 (sparse)}&\textbf{2.35}$^+$(3\%)&\textbf{9.36}$^+$(3\%)& 1.35(4\%) &7.83$^+$(4\%)&3.08(5\%)&5.22(6\%)&\textbf{5.36}(3\%)&\textbf{3.83}(4\%)&\textbf{1.7}(6\%)\\
\textbf{MTS2.0 (semi-dense)}&6.51(24\%)&\---&$2.24$(25\%)&\---&3.72(20\%)&7.04(19\%)&7.38(15\%)&6.54(25\%)&2.61(30\%) \\
MTS1.0 (sparse)&5.47$^+$(2\%) &15.5$^+$(2\%)&1.58(2\%)&8.92$^+$(3\%)&2.48(2\%)&7.32(2\%)& 8.8(1\%)&7.22(3\%)&3.03(4\%)\\
MTS1.0 (semi-dense)&17.47(57\%) &\---&4.47(44\%)&\---& 3.78(18\%)&12.8(14\%)&16.2(38\%)&15.41(40\%)&6.93(58\%)\\
SGBM1 \cite{Hirschmuller2008}&7.83$^+$(67\%)&16.3$^+$(63\%)&1.45(84\%)&\---&2.61(89\%)&4.77(92\%)&16.06(70\%)&11.37(81\%)&5.14(86\%)\\
SGBM2 \cite{Hirschmuller2008}&8.92$^{+*}$(83\%)&15.9$^{+*}$(77\%)&\textbf{1.27}(82\%)&\textbf{5.86}$^+$(90\%)&2.16(90\%)&\textbf{4.67}(90\%)&15.72(64\%)&10.5(78\%)&4.58(85\%)\\
ELAS\_ROB \cite{geiger2010efficient} &10.5$^{+*}$&13.4$^{+*}$&1.49(99\%)&9.67$^+$&\textbf{2.06}&7.02&11.71&17.75&7.46\\
FPGA Stereo \cite{honegger2017embedded}&\---&\---&\---&\---&2.94\cite{pu2019sdf}&11.41\cite{pu2019sdf}&\---&\---&\---\\
 \hline 
DispNet \cite{mayer2016large}&\---&\---&\textbf{0.68}\cite{mayer2016large}&4.34$^+$&\textbf{1.35}\cite{pu2019sdf}&\textbf{6.28}\cite{pu2019sdf}&\textbf{15.62}$^*$\cite{mayer2016large}&\textbf{5.78}\cite{mayer2016large}&1.68\cite{mayer2016large} \\
EdgeStereo \cite{song2019edgestereo}&\textbf{2.00}$^+$&\textbf{3.72}$^+$&2.07(?\%)\cite{song2019edgestereo}&\textbf{2.08}$^+$&\--- &\---&\---&\---&\textbf{0.74}$^{*T}$(?\%)\cite{song2019edgestereo}\\
iResNet \cite{liang2017learning}&\---&\---&1.21(?\%)\cite{song2019edgestereo}&2.44$^T$(?\%)\cite{song2019edgestereo}&\---&\---&\---&\---&0.95$^{*T}$(?\%)\cite{song2019edgestereo}\\
     \bottomrule
  \end{tabular}
\end{table*}

\begin{table*}[!h]
  \centering
  \setlength{\tabcolsep}{3pt}
  \renewcommand{\arraystretch}{1.3}  
   \caption{Results on the Middlebury training benchmark expressed in \textit{avgerr}. Results of MTStereo 2.0 are bold. All results taken from respective benchmark website, except that of Semi-Dense. We mark with $^*$ the results that were computed on a different set of images than we used for evaluation.}
  \label{tab:accuracyResultsMiddlebury}
   \begin{tabular}{cccccccccccc}
    \toprule
   MotionStereo \cite{valentin2018depth}&\textbf{Sparse} &SNCC \cite{einecke2010two}&LS-ELAS \cite{jellal2017ls}&ELAS \cite{geiger2010efficient}&SED \cite{pena2016disparity}&\textbf{Semi-Dense} &SGBM1 \cite{Hirschmuller2008}&SGBM2 \cite{Hirschmuller2008} \\

 \cmidrule(lr){1-1}  \cmidrule(lr){2-2}     \cmidrule(lr){3-3}     \cmidrule(lr){4-4}     \cmidrule(lr){5-5}     \cmidrule(lr){6-6}     \cmidrule(lr){7-7}     \cmidrule(lr){8-8} 	 \cmidrule(lr){9-9} 

  1.72$^*$(46\%)&2.35(3\%)&3.25$^*$(62\%)&4.35(61\%)&4.94(73\%)&5.38(1\%)&6.51(24\%)&7.83(67\%)&8.92$^*$(83\%)\\
     \bottomrule
  \end{tabular}
\end{table*}

\begin{table*}[!h]
  \centering
  \renewcommand{\arraystretch}{1.3}  
   \caption{Results on the Middlebury test benchmark expressed in \textit{avgerr}. Results of MTStereo 2.0 are bold. All results taken from respective benchmark website. We mark with $^*$ the results that were computed on a different set of images than we used for evaluation.}
  \label{tab:accuracyResultsMiddleburyTest}
   \begin{tabular}{ccccccccccc}
    \toprule
   MotionStereo \cite{valentin2018depth} &SNCC \cite{einecke2010two}&LS-ELAS \cite{jellal2017ls} &\textbf{Sparse} &ELAS \cite{geiger2010efficient}&SED \cite{pena2016disparity}&SGBM1 \cite{Hirschmuller2008}&SGBM2 \cite{Hirschmuller2008}   \\

 \cmidrule(lr){1-1}  \cmidrule(lr){2-2}     \cmidrule(lr){3-3}     \cmidrule(lr){4-4}     \cmidrule(lr){5-5}     \cmidrule(lr){6-6}     \cmidrule(lr){7-7}     \cmidrule(lr){8-8} 	  

 3.30$^*$(38\%)  &3.96$^*$(55\%) &9.10(49\%) &9.36(3\%) &10.6(66\%) &12.3(2\%) &16.3(63\%) &15.9$^*$(77\%)  \\
     \bottomrule
  \end{tabular}
\end{table*}

\subsection{Evaluation}
We computed standard metrics for the concerned benchmarks, allowing direct comparison with existing methods:
\begin{itemize}
    \item \textbf{avgerr}: the average absolute disparity error (in pixels) among all pixels of which both a disparity value was in the ground truth and estimated disparity map.
    \item \textbf{D-all-est}: the percentage of stereo disparity outliers (the disparity error is $\geq$3px and $\geq$5\% of the true disparity) among all pixels of which both a disparity value was in the ground truth and estimated disparity map.
    \item \textbf{Density}: the percentage of pixels with a disparity prediction with respect to the total number of pixels in the reference image. Values were rounded to the nearest decimal.
    \item \textbf{time/MP}: the execution time in seconds normalized by the number of megapixels in the reference image.
\end{itemize}

We defined a single set of parameters that contributes to achieve robust performance across the different benchmark data sets. The number of color quantization levels $q$ was set to 16 (8) when sparse (semi-dense) disparity maps were generated. The weight $\alpha$ of the context cost relative to that of the gradient cost was set to 0.8. The minimum (or maximum) width of nodes to be matched $\omega_\alpha$ (or $\omega_\beta$) was set to 0 (or $1/2$ of the input image width). Matched node levels $S$ was set to $\{1, 0\}$. The maximum neighbourhood size $\omega_\gamma$ was set to 10. The size of the Gaussian kernel used to aggregate the cost volume $\omega_{cv}$ was $21$. The minimum confidence percentage parameter $\omega_\Pi$ was set to $12$ during coarse-to-fine matching. In guided pixel refinement, $\omega_\Pi$ was set to $12\%$ ($4\%$) when sparse (semi-dense) disparity maps were generated. $\omega_\omega$ was set to $15\%$.

\subsection{Results and comparison}

In Table~\ref{tab:accuracyResults}, we report the results achieved by our method on the considered data sets as well as those of other existing methods that do not formulate the stereo matching as a learning problem (upper part) and those that deploy neural networks to tackle the disparity estimation problem (lower part). The results of MTStereo 1.0 were generated using the parameters used in~\cite{BRANDT2020}. We set the parameter $\omega_\beta = \frac{1}{3}$ in all experiments except for those involving the Trimbot datasets, wherein we set $\omega_\beta = \frac{1}{15}$.
In Table~\ref{tab:accuracyResultsMiddlebury} and Table~\ref{tab:accuracyResultsMiddleburyTest}, we report the error achieved by our method on the Middlebury training and test data, respectively, as well as those of other existing methods that have low average time/MP and do not run on a GPU.
The lower the values, the better the performance.
%
Both the sparse and semi-dense versions of our method achieved better or comparable accuracy results than those of many existing methods.
The accuracy results that we achieved, especially in the case of the sparse estimation, are comparable to those achieved by approaches that deploy architectures based on convolutional networks and are trained with large amount of labeled data. 

The results of MTStereo 2.0 sparse and semi-dense are generally more accurate than those of their MTStereo 1.0 counterpart method. Note that the density of the disparity maps produced by the methods are influenced by the parameters, which can be used to trade-off between accuracy and density. The cases where MTStereo 2.0 produces more accurate and dense disparity maps than MTStereo 1.0 (e.g. for the sparse variants when evaluated on the Middlebury, Kitti2015, Synth garden, and Driving data sets) suggest that MTStereo 2.0 is a more robust method than MTStereo 1.0. 

The average disparity error (avgerr) of disparity maps produced by MTStereo 2.0 sparse is often lower than that of disparity maps produced by the other listed methods. The results on the Middlebury data sets show, for example, the avgerr in sparse disparity maps produced by MTStereo 2.0 is lower than that of all other listed methods except MotionStereo. Also, the avgerr on the Kitti training data set of MTStereo 2.0 sparse is lower than that of all other methods except DispNet, iResNet, and SGBM2. Furthermore, the avgerr of disparity maps produced by MTStereo 2.0 sparse is lower than all other non-learning based methods when evaluated on Driving, Monkaa, and Flying Things 3D. MTStereo 2.0 sparse produced more accurate disparity maps than some of the listed CNN based methods when evaluated on the Kitti2015 training, Synthetic garden, Driving, and Monkaa data sets. Note that the performance of CNN based methods depend on the used training set, and the densities of the disparity maps produced by said CNN based methods are higher.
The avgerr of disparity maps produced by MTStereo 2.0 semi-dense is often lower than that of disparity maps produced by other listed methods. The averr achieved by MTStereo 2.0 is frequently close to that of MTStereo 1.0 sparse, while it produces more dense disparity maps. 

The application of stereo matching in robotics introduces challenges: robots frequently have a tight power budget, real-time performance is required, and stereo cameras can be of low quality. The Real Garden and Synthetic Garden data sets contain stereo pairs that are observed by the TrimBot2020 gardening robot~\cite{StrisciuglioTrimbot}. The performance of our method on the Real Garden and Synthetic Garden data sets, which contains low resolution stereo pairs, is comparable to that of the existing non-learning based methods. Together with the implementation of MTStereo 2.0, we released a ROS-node version optimized for scenes containing plants, and examples of point clouds it produces, as the algorithm was deployed on the TrimBot2020 robot\footnote{https://www.youtube.com/watch?v=ZX3rPrq6x5g}. 
The accuracy of disparity maps produced by learning based methods is affected by differences between the data set used for training and that used for evaluation. The results of DispNet on Kitti2015 were obtained with a model trained on the FlyingThings3D data set, and subsequently fine-tuned on the Kitti2015 data set. When DispNet is not fine-tuned on Kitti2015, its error (avgerr) increases to 1.59 \cite{mayer2016large}. The different characteristics of the data sets seem to limit performance.

\captionsetup[subfigure]{labelformat=empty}
\begin{figure*}[!t]
\centering
\setlength{\unitlength}{38mm}
\hspace{20pt}\makebox[\unitlength][c]{Example image}
\makebox[\unitlength][c]{Ground truth}
\makebox[\unitlength][c]{Semi-Dense}
\makebox[\unitlength][c]{Sparse}\\ \vspace{4pt}
\makebox[20pt]{\raisebox{30pt}{\rotatebox[origin=c]{90}{Middlebury}}}%
\includegraphics[width=\unitlength]{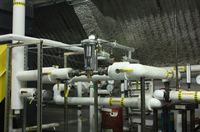}
\includegraphics[width=\unitlength]{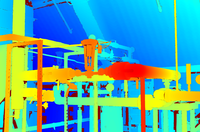}	\includegraphics[width=\unitlength]{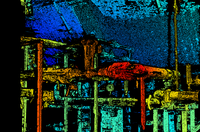}
\includegraphics[width=\unitlength]{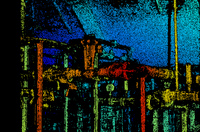}\\ \vspace{4pt}
\makebox[20pt]{\raisebox{30pt}{\rotatebox[origin=c]{90}{Synth Garden}}}%
\includegraphics[width=\unitlength]{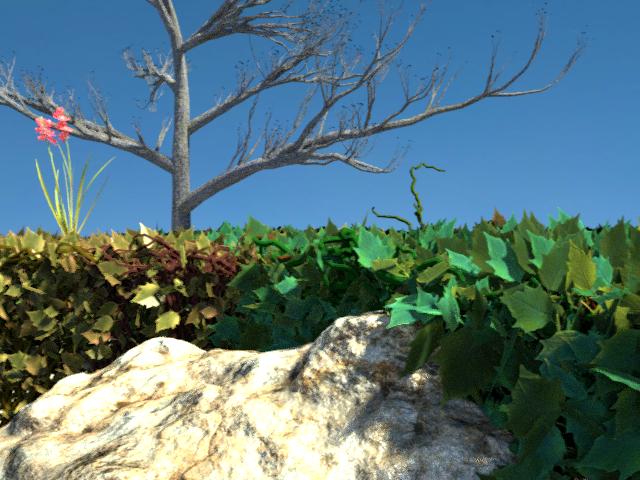}
\includegraphics[width=\unitlength]{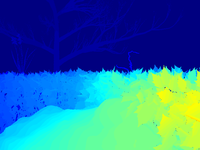}
\includegraphics[width=\unitlength]{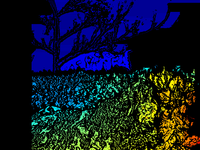}	\includegraphics[width=\unitlength]{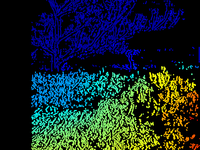}\\ \vspace{4pt}
\makebox[20pt]{\raisebox{30pt}{\rotatebox[origin=c]{90}{Monkaa}}}%
\includegraphics[width=\unitlength]{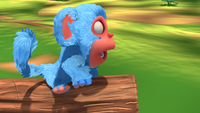}
\includegraphics[width=\unitlength]{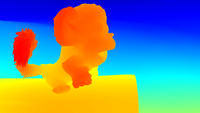}
\includegraphics[width=\unitlength]{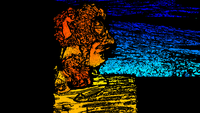}
\includegraphics[width=\unitlength]{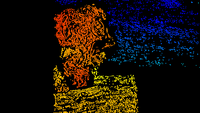}\\ \vspace{4pt}
\makebox[20pt]{\raisebox{30pt}{\rotatebox[origin=c]{90}{Flying things}}}%
\includegraphics[width=\unitlength]{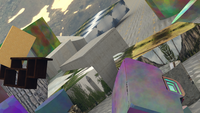}
\includegraphics[width=\unitlength]{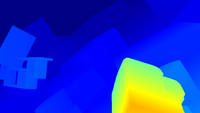}
\includegraphics[width=\unitlength]{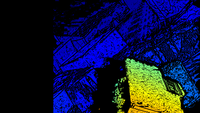}
\includegraphics[width=\unitlength]{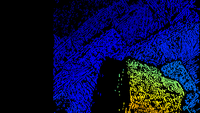}
	
	\caption{Example images from the Middleburry, Trimbot2020 Synthetic Garden, Monkaa, and Flying Things 3D data sets, with corresponding ground truth depth images, our semi-dense and sparse reconstruction. Morphological dilatation was applied to sparse outputs for visualization purposes.}
	\label{fig:examples}
\end{figure*}

\captionsetup[subfigure]{labelformat=empty}
\begin{figure*}[!t]
\centering
\setlength{\unitlength}{50mm}\makebox[\unitlength][c]{Example image}
\makebox[\unitlength][c]{Error MTStereo v1}
\makebox[\unitlength][c]{Error MTStereo v2}\\ \vspace{4pt}
\makebox[20pt]{\raisebox{20pt}{\rotatebox[origin=c]{90}{Semi-Dense}}}%
\includegraphics[width=\unitlength]{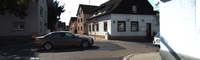}
\includegraphics[width=\unitlength]{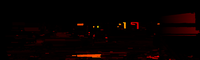}	\includegraphics[width=\unitlength]{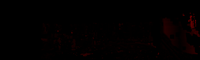}\\
\makebox[20pt]{\raisebox{20pt}{\rotatebox[origin=c]{90}{Sparse}}}%
\makebox[\unitlength][c]{}
\includegraphics[width=\unitlength]{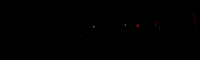}	\includegraphics[width=\unitlength]{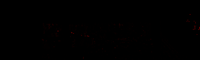}\\
	\caption{Difference between the accuracy of MTStereo 1.0 and 2.0. Example image taken from kitti2015 training data set. Morphological  dilatation  was  applied to all outputs for visualization purposes.}
	\label{fig:v1vsv2}
\end{figure*}

In Fig.~\ref{fig:examples}, we show example images from the Middlebury, Trimbot2020 Synthetic Garden, Monkaa, and Flying Things 3D data sets, with corresponding ground truth depth images, our semi-dense and sparse depth reconstruction.
One can notice that our method is able to robustly extract depth information also in image regions with little texture. For example, the depth of the gray cubes in the Flying Things example image is reasonably well estimated although the cube surface are rendered with plain colors.  
Regions with very little or no texture have usually an inherent disparity ambiguity, and our method allows for controlling such cases. When the parameter $\omega_\Pi$ is set to a low value, an assumption is made that regions with little texture are flat. The dis-ambiguous disparity values at the left and right side edges of regions with little or no texture are then linearly interpolated within the region. 
When the assumption is not correct, such as between the tree branches in the MTStereo semi-dense disparity map of the Synth Garden example, the parameter $\omega_\Pi$ can be increased such that no disparity values are assigned in the ambiguous region. 
Edge blurring artifacts are largely not present, as it can be seen for the edges of the pipes in the disparity maps of the Middlebury example (Fig.~\ref{fig:examples}, first row) or those of the tree in the estimated maps of the Synth Garden esample (Fig.~\ref{fig:examples}, second row). The disparity maps produced by our method, although sparse, are accurate and contain (very) little outliers. However, the estimated maps are dense enough to support applications such as robot navigation or visual servoing.

In Fig.~\ref{fig:v1vsv2}, we show the difference between the accuracy of MTStereo 1.0 and 2.0 (lighter pixels have greater error). MTStereo 2.0 is able to more robustly extract depth information than MTStereo 1.0. The disparity maps produced by MTStereo 2.0 generally contain less estimations with great error than those produced by MTStereo 1.0. 

The time in seconds needed to process image pairs normalized by their size in megapixels (time/MP) was for sparse (semi-dense) estimation by MTStereo v1 on Middlebury training 0.54 (0.52), Kitti2015 training 0.39 (0.36), Real Garden 0.37 (0.32), Synth Garden 0.25 (0.22), Driving 0.32 (0.29), Monkaa 0.28 (0.24), and Flying Things 0.34 (0.30). 
The time in seconds needed to process image pairs normalized by their size in megapixels (time/MP) was for sparse (semi-dense) estimation by MTStereo v2 on Middlebury training 7.43 (7.23), Kitti2015 training 4.45 (4.05), Real Garden 1.76 (1.93), Synth Garden 1.72 (1.88), Driving 4.02 (3.21), Monkaa 2.67 (3.21), and Flying Things 3.88 (3.95).

The processing times of methods included in this paper are listed on the Middlebury and Kitti benchmark websites, and are generally lower than those needed by MTStereo 2.0. The efficiency of our code can be increased through code-level optimization, e.g. of our cost volume implementation. The average relative time needed to process the different steps of MTStereo 2.0 on Kitti2015 training was for sparse estimation 24\% computing the cost volume, 20\% coarse node determination, 45\% coarse to fine matching (of which 22\% cost volume lookup), 7\% noise filtering, and the remaining processing time was divided over the other steps. For semi-dense estimation 27\% computing the cost volume, 12\% coarse node determination, 40\% coarse to fine matching (of which 31\% cost volume lookup), 17\% noise filtering, and the remaining processing time was divided over the other steps. A significant portion of the processing time is hence consumed by cost volume related tasks. 


\section{CONCLUSIONS}
We proposed a stereo matching method, called MTStereo 2.0, for systems with limited computational resources that require efficient and accurate depth estimation. 
It improves on its predecessor MTStereo 1.0 as it:
\begin{inparaenum}[\itshape a\upshape)]
    \item deploys a more robust cost function,
    \item performs more thorough detection of incorrect matches,
    \item computes disparity maps with pixel-level rather than node-level precision.
\end{inparaenum}

MTStereo 2.0 produces disparity maps which are generally more accurate than those produced by MTStereo 1.0, and does not require intensive GPU computations like methods based on CNNs. It can thus run on embedded and robotics devices with low-power requirements. The higher accuracy achieved by MTStereo 2.0 w.r.t. its predecessor is attributable to the thorough re-design that we made. 

Our method achieves competitive results on several benchmark data sets:  Middlebury 2014, KITTI 2015, Driving, FlyingThings3D, Monkaa and the TrimBot2020 garden. The density of the MTStereo 2.0 disparity maps is enough for many applications of robot navigation or visual servoing. This is demonstrated in the implementation that we release, together with its ROS version, at \url{https://github.com/rbrandt1/MaxTreeS}.










\section*{ACKNOWLEDGMENTS}
This work received support from the EU Horizon 2020 program, under the project TrimBot2020 (grant No. 688007)

\bibliographystyle{bibliography/IEEEtran}
\bibliography{bibliography/IEEEabrv,refs}

\end{document}